\documentclass[10pt,twocolumn,letterpaper]{article}

\usepackage[pagenumbers]{cvpr}  %

\usepackage{graphicx}
\usepackage{amsmath}
\usepackage{amssymb}
\usepackage{booktabs}
\usepackage{caption}
\usepackage{xcolor, eucal}
\newcommand{\myparagraph}[1]{{\vspace{.1em} \noindent \bf #1}}
\def\Ours{{Free SOLO}\xspace}
\def\Ours{{FreeSOLO}\xspace}

\renewcommand{\texttt}[1]{{{$\tt #1$}}}

\usepackage{pifont}%
\usepackage{url}            %
\usepackage{bbm,nicefrac}
\usepackage{bm}
\newcommand{\app}{\raise.17ex\hbox{$\scriptstyle\sim$}}
\newcommand{\tablestyle}[2]{\setlength{\tabcolsep}{#1}\renewcommand{\arraystretch}{#2}\centering\footnotesize}
\usepackage{makecell, verbatim, sidecap}
\usepackage{multirow}
\usepackage{enumitem}
\usepackage[accsupp]{axessibility}  %

\usepackage[pagebackref=true,breaklinks=true,letterpaper=true,colorlinks,urlcolor=black,bookmarks=false]{hyperref}
\hypersetup{citecolor=[RGB]{119,185,0}}

\renewcommand{\vec}[1]{\ensuremath{\pmb{#1}}}
\newcommand{\mat}[1]{\ensuremath{\mathbf{#1}}}
\newcommand{\set}[1]{\ensuremath{\mathscr{#1}}}
\makeatletter
\@tfor\next:=abcdefghijklmnopqrstuvwxyxz\do
 {\begingroup\edef\x{\endgroup
    \noexpand\@namedef{v\next}{\noexpand\vec{\next}}%
  }\x}
\@tfor\next:=ABCDEFGHIJKLMNOPQRSTUVWXYZ\do
 {\begingroup\edef\x{\endgroup
    \noexpand\@namedef{m\next}{\noexpand\mat{\next}}%
  }\x}
\@tfor\next:=ABCDEFGHIJKLMNOPQRSTUVWXYZ\do
 {\begingroup\edef\x{\endgroup
    \noexpand\@namedef{s\next}{\noexpand\set{\next}}%
  }\x}
\makeatother

\makeatletter
\let\UrlSpecialsOld\UrlSpecials
\def\UrlSpecials{\UrlSpecialsOld\do\/{\Url@slash}\do\_{\Url@underscore}}%
\def\Url@slash{\@ifnextchar/{\kern-.11em\mathchar47\kern-.2em}%
    {\kern-.0em\mathchar47\kern-.08em\penalty\UrlBigBreakPenalty}}
\def\Url@underscore{\nfss@text{\leavevmode \kern.06em\vbox{\hrule\@width.3em}}}
\makeatother

\usepackage[capitalize]{cleveref}
\Crefname{section}{Section}{Sections}
\crefname{section}{Sec.}{Secs.}
\Crefname{table}{Table}{Tables}
\crefname{table}{Table}{Tables}

\def\cvprPaperID{1233} %
\def\confName{CVPR}
\def\confYear{2022}

\newcommand{\sd}[1]{}

\begin{document}

\title{FreeSOLO: Learning to Segment Objects without Annotations\thanks{Part of this work was done when XW was an intern at NVIDIA, and CS was with The Univerity of Adelaide. CS is the corresponding author.}
}

\author{
Xinlong Wang$^1$, ~~~
Zhiding Yu$^2$, ~~~
Shalini De Mello$^2$, ~~~
Jan Kautz$^2$, \\
Anima Anandkumar$^{2,3}$, ~~~
Chunhua Shen$^4$,  ~~~
Jose M. Alvarez$^2$ \\[0.25cm]
$^1$ The University of Adelaide 
~~~~~
$^2$ NVIDIA
~~~~~
$^3$ Caltech
~~~~~ $^4$ Zhejiang University
}

\makeatletter
\let\@oldmaketitle\@maketitle%
\renewcommand{\@maketitle}{\@oldmaketitle%
\centering
    \vspace{-0.21cm}
    \includegraphics[trim=0 .25cm 0 0,clip,width=1.0\linewidth]{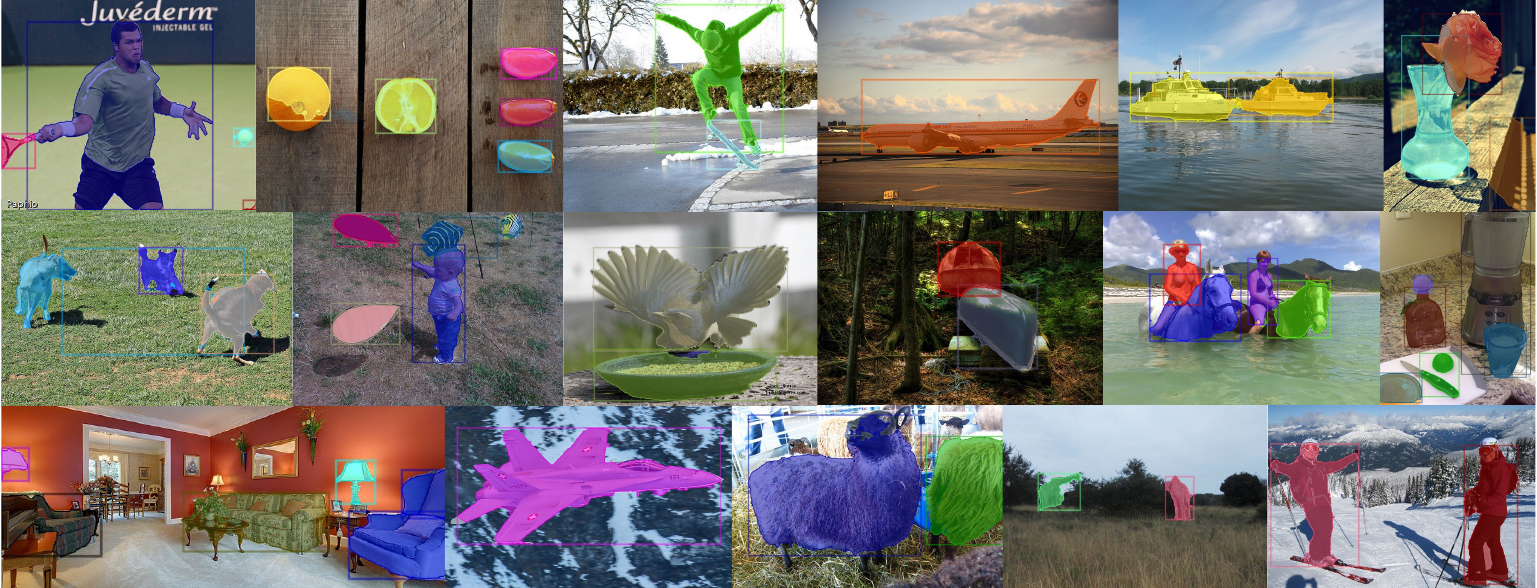}
    \vspace{-0.55cm}
    \captionof{figure}{%
    \textbf{Qualitative results of \Ours
    for the task of class-agnostic instance segmentation.}
    The model is trained \textit{without any kind of manual annotations} and can infer at 16 FPS on a V100 GPU.  
    Best viewed on screen.
    }
    \label{fig:vis_first}
    \bigskip}                   %
\makeatother

\maketitle

\begin{abstract}
\vspace{-1em}
Instance segmentation is a fundamental vision task that aims to recognize and segment each object in an image.
However, it requires costly annotations such as bounding boxes and segmentation masks for learning.
In this work, we propose a fully unsupervised learning method that learns class-agnostic instance segmentation without any annotations.
We present \Ours, a self-supervised instance segmentation framework built on top of the simple instance segmentation method SOLO.
Our method also presents a novel localization-aware pre-training framework, where objects can be discovered from complicated scenes in an unsupervised manner. 
\Ours achieves 9.8\% AP$_{50}$ on the challenging COCO dataset, which even outperforms several segmentation proposal methods that use manual annotations.
For the first time, we demonstrate unsupervised class-agnostic instance segmentation successfully.
\Ours's box localization significantly outperforms state-of-the-art unsupervised object detection/discovery methods, with about 100\% relative 
improvements in COCO AP.
\Ours further demonstrates superiority as a strong pre-training method, outperforming state-of-the-art self-supervised pre-training methods by $+$9.8\% AP when fine-tuning instance segmentation with only 5\% COCO masks.

Code is available at: 
 \href{https://github.com/NVlabs/FreeSOLO}{\sf\rm\small\ttfamily\color{magenta} github.com/NVlabs/FreeSOLO}

\end{abstract}
\vspace{-6.5mm}

\section{Introduction}

Instance segmentation is a fundamental computer vision task that requires recognizing the objects in an image and segmenting each of them at the pixel level. 
Instance segmentation subsumes object detection, 
as bounding box can be thought of as a coarse parametric representation of a segmentation mask. 
Therefore, it is a more demanding and challenging task than object detection by requiring both instance-level and pixel-level predictions. Recently, significant progress~\cite{fcis, he2017mask, de2017semantic, chen2019hybrid, yolact, CondInst, wang2021SOLO} has been made to address the instance segmentation task. However, the dense prediction nature of the task  requires rich and expensive annotations during training. Weakly-supervised instance segmentation methods are thus proposed to relax the annotation requirements~\cite{KhorevaBH0S17,hsu2019weakly,liu2020leveraging,tian2020boxinst,cheng2021pointly,lan2021discobox}. Latest methods such as BoxInst~\cite{tian2020boxinst} and DiscoBox~\cite{lan2021discobox} have significantly closed the gap to fully supervised methods. However, their competitive result still relies on box or point annotations that contain strong localization information.

\begin{figure*}[t!]
\centering
\includegraphics[width=0.90859095\linewidth]
{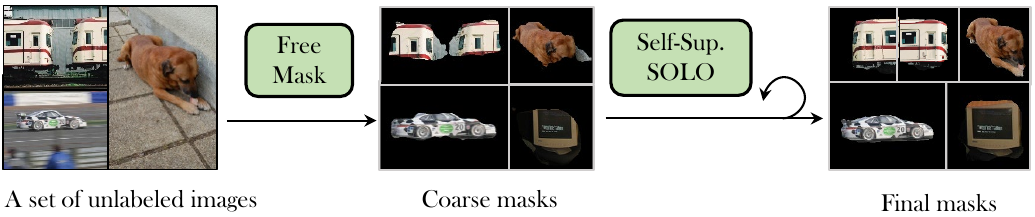}
\vspace{-0.5em}
\caption{\textbf{Overview of \Ours{}}. Unlabeled images are first input to Free Mask to generate coarse object masks. The segmentation masks as well as their associated semantic embeddings are used to train a SOLO-based instance segmentation model via weak supervision. We use self-training to improve object mask segmentation.
} 
\label{fig:free_solo_pipeline}
\vspace{-1.0em}
\end{figure*}

In this work, we 
explore \textit{learning class-agnostic instance segmentation without any annotations.} 
The work here is built upon our recent work of 
SOLO~\cite{wang2021SOLO}, a simple %
yet 
strong instance segmentation framework, and the 
self-supervised dense feature learning method of 
DenseCL \cite{wang2020DenseCL}. 
SOLO adopts a one-stage design, which contains a category branch and a mask branch to encode the object category information and segmentation proposals, respectively. Our main intuition is that this ``top-down meets bottom-up'' design allows us to unify pixel grouping, object localization and feature pre-training in a fully self-supervised manner. 

Our proposed framework, \textbf{\Ours}, contains two major pillars: Free Mask and Self-supervised SOLO, as shown in Figure~\ref{fig:free_solo_pipeline}. 
Specifically, Free Mask contains self-supervised  design elements that promote objectness in network attention. 
It contains a ``query-key'' attention design, 
where the queries and keys are constructed from self-supervised features.
The method takes the cosine similarity between each query with all the keys, thus obtaining a set of query-conditioned (seeded) attention maps as coarse masks. The coarse masks are ranked and filtered by their maskness scores, followed by non-maximum suppression (NMS) to further remove the redundant masks. Self-Supervised SOLO then takes the coarse masks as pseudo-labels to train a SOLO model. Since the coarse masks can be inaccurate, Self-Supervised SOLO contains a weakly-supervised design to better accommodate the label noise. 
This is followed by a self-training strategy to further refine mask quality and to improve accuracy. 
Our network design is almost the same as SOLO with minimal modifications, thus leading to simple and fast  inference process.

\Ours provides an effective solution to the challenging problem of self-supervised instance segmentation. With the bounding boxes obtained from the predicted masks, \Ours also shows significant advantage as an unsupervised object discovery method. In addition to the above roles, we further consider FreeSOLO as a strong self-supervised pretext task for instance segmentation by jointly learning object-level and pixel-level representations.
Compared to pre-training for image classification~\cite{moco, simclr, byol}, object detection~\cite{updetr, detcon2021} and semantic segmentation~\cite{PinheiroABGC20, ChaitanyaEKK20}, pre-training for instance segmentation is still under-studied. General instance segmentation requires not only localizing objects at the pixel level, but also recognizing their semantic categories. 
Interestingly, the design of FreeSOLO allows us to directly learn object-level semantic representations in an unsupervised manner. 
Upon completing the pre-training, all the learned parameters except for the last classification layer can be used to initialize the supervised instance segmentation models to improve accuracy.

\noindent 
Our contributions can be summarized as follows.

\begin{itemize}[leftmargin=*,nosep]

\item

We propose the Free Mask approach, which leverages the specific design of SOLO to effectively extract coarse object masks and semantic embeddings in an unsupervised  manner.

\item
We further propose Self-Supervised SOLO, which takes the coarse masks and semantic embeddings from Free Mask and trains the SOLO instance segmentation model, with several novel design elements to overcome label noise in the coarse masks.

\item 
With the above methods, \Ours presents a simple and effective framework that demonstrates unsupervised  instance segmentation successfully for the first time.
Notably, it outperforms some proposal generation methods that use manual annotations. \Ours also outperforms state-of-the-art methods for unsupervised object detection/discovery by a significant margin (relative +100\% in COCO AP). 

\item
In addition, \Ours serves as a strong self-supervised pretext task for representation learning for instance segmentation. 
For example, when fine-tuning on  COCO dataset with 5\% labeled masks, \Ours outperforms DenseCL~\cite{wang2020DenseCL} by +9.8\%  AP.
	
\end{itemize}

\section{Related Work}
\vspace{-0.5em}

\myparagraph{Instance segmentation.} 
Instance segmentation has attracted much attention in recent years.
Most existing works focus on learning instance segmentation with full annotations.
Top-down methods~\cite{fcis, he2017mask, panet, chen2019hybrid}
solve the problem from the perspective of object detection, \ie, detecting the bounding box of objects first and then segmenting the object in the box.
Bottom-up methods~\cite{associativeembedding, de2017semantic, SGN17, Gao_2019_ICCV} view the task as a label-then-cluster problem, \eg, by learning per-pixel embeddings first and then clustering them into groups.
Some recent methods~\cite{yolact, chen2020blendmask, CondInst, wang2020solo, wang2020solov2} seek a combination of top-down and bottom-up approaches to perform faster inference and better segmentation.
Among these methods, SOLO has shown a promising speed/accuracy trade-off with a very simple architecture.
A few works explore learning instance segmentation with weak annotations, \eg, image-level and box-level labels~\cite{KhorevaBH0S17, Zhou2018PRM, hsu2019weakly, tian2020boxinst}. To the best of our knowledge, none have additionally explored learning instance segmentation without any labels at all.

In particular, 
BoxInst \cite{tian2020boxinst} attains strong instance segmentation results using box annotations only, demonstrating that instance segmentation may not necessarily be more difficult to solve than box-level object detection. We move  one-step forward 
by reporting strong instance segmentation results in an unsupervised setting,  without any annotations.  

\myparagraph{Self-supervised learning.} 
To learn a good visual representation from unlabeled data, a wide range of pretext tasks have been explored, \eg, colorization~\cite{zhang2016colorful}, inpainting~\cite{inpainting16}, jigsaw puzzles~\cite{jigsaw} and orientation discrimination~\cite{gidaris2018rotations}.
The breakthroughs came from the contrastive learning methods, \eg, SimCLR \cite{simclr} and MoCo~\cite{moco} that perform an instance discrimination pretext task~\cite{wu2018unsupervised}.
Besides pre-training for image classification~\cite{byol, swav20, simsiam21}, some recent works~\cite{wang2020DenseCL, xie2020propagate, detcon2021, xie2021detco, xiao2021region} design self-supervised pre-training methods for dense prediction tasks, \eg, object detection and semantic segmentation.
Different from them, our method can not only learn intermediate representations, but also train instance segmenters, which can segment objects in the wild. 
Our \Ours naturally serves as a strong pretext task for learning representations for instance segmentation. 
The pre-trained model can be seamlessly transferred to supervised fine-tuning and can achieve significant gains compared to existing pre-training methods.

\myparagraph{Unsupervised object discovery.} 
A wide range of approaches have been proposed for unsupervised object discovery, 
including statistical topic discovery models~\cite{Sivic2005DiscoveringOC, RussellFESZ06},  link analysis technique~\cite{KimT09}, clustering by composition~\cite{FaktorI14}, and part-based matching~\cite{ChoKSP15}.
Some recent works~\cite{VoBCHLPP19, vo2020toward} formulate object discovery as an optimization problem.
LOD~\cite{lod} further proposes to formulate unsupervised object discovery as a ranking problem.
Yet, the existing methods have achieved limited success in challenging and complicated scenes.
Furthermore, most of these methods can only find coarse bounding boxes of objects.
By contrast, our method discovers and localizes objects in the wild with pixel-wise segmentation masks.
With bounding boxes obtained from predicted masks,  
\Ours  outperforms the state-of-the-art unsupervised object discovery methods by a large margin.

\myparagraph{Unsupervised segmentation.}
To remove the dependency on manual supervision, some object co-segmentation works~\cite{JoulinBP10, HsuLC18, ChenL0H21} make a strong assumption  about the image collection, \ie, to segment common repeated objects in a collection of images.
Besides, there are a few works~\cite{JiVH19, HwangYSCYZC19, maskcontrast} that explore unsupervised semantic segmentation.
Some~\cite{JiVH19} only deal with simple scenarios, and some~\cite{HwangYSCYZC19, maskcontrast} still require a salient object estimator or boundary annotations.
In addition, the key difference lies in the task.
Instead of semantic segmentation, our method solves the harder problem of instance segmentation, \ie, to segment each object individually.

\section{Method}
\vspace{-0.5em}

\myparagraph{Background.}
We briefly introduce the supervised instance segmentation method SOLO~\cite{wang2021SOLO}.
SOLO %
shows
that instance segmentation can be solved by directly mapping an input image to the desired object categories and instance masks %
using 
fully convolutional
networks (FCNs), eliminating the need for bounding box detection or grouping via post-processing.
Its main idea is to formulate instance segmentation %
into 
two simultaneous category-aware pixel-level prediction problems.
It conceptually divides the input image into $S \times S$ grids. 
A grid cell is responsible for predicting the semantic category as well as the segmentation mask for an object whose center falls into that grid cell.
The model consists of two branches, \ie, a category branch and a mask branch.
The category branch predicts the semantic categories.
The mask branch generates $S^2$ sized masks, one corresponding to each grid cell.
Specifically, the dynamic SOLO variant employs dynamic convolutions to separately predict the mask kernels and mask features. The mask features are then convolved with the predicted mask kernels to generate the masks. 
This operation can be written as:
\begin{equation}
\begin{aligned}
\mS = {\mG} \circledast {\mF},
\end{aligned}
\end{equation}
where $\mG$ is the convolution kernel, and $\mS$ denotes the score maps for all the $S^2$ masks.
$\mS$ is then normalized via a $\tt sigmoid$ operation, and input to mask NMS to form the final object masks.

\subsection{Overview of \Ours}
\label{sub:pipeline}

We propose a novel framework for self-supervised instance segmentation, termed \Ours.
\Ours does not require any type of annotations, neither pixel-level nor image-level labels, and simply uses a collection of unlabeled images for training.
Its overall pipeline is illustrated in Figure~\ref{fig:free_solo_pipeline}.
We first propose the Free Mask approach to generate segmentation masks from a self-supervised pre-trained model.
For each unlabeled image, the coarse object masks can be generated fast with simple operations, \eg, at 21 FPS on a V100 GPU with a ResNet-50-based backbone.
We further propose Self-Supervised SOLO, which trains the SOLO-based instance segmenter using the coarse masks and semantic embeddings from Free Mask, with several novel design elements including weaky-supervised design, self-training, and semantic embedding learning.

With \Ours, we obtain an instance segmentation model given only unlabeled images.
In addition to unsupervised instance segmentation itself, the well-trained model serves as a strong pre-trained  model for downstream fine-tuning.
All its parameters except the last classification layer can be transferred to supervised instance segmentation as a strong initialization.

\subsection{Free Mask}
\label{sub:freemask}

\def\SM{\mathbf{\Delta}}
\def\vd{\mathbf{d}}
\def\mF{\mathbf{F}}
\def\mG{\mathbf{G}}
\def\mQ{\mathbf{Q}}
\def\mK{\mathbf{K}}
\def\mI{\mathbf{I}}
\def\mM{\mathbf{M}}
\def\mS{\mathbf{S}}
\def\mD{\mathbf{\Theta}}

\begin{figure}[t!]
\centering
\includegraphics[width=1.0\linewidth]{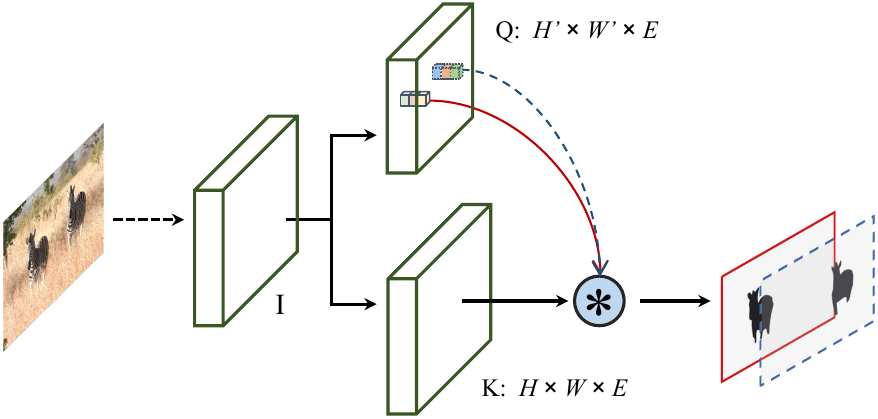}
\vspace{-.8em}
\caption{
\textbf{The 
Free Mask approach.} 
Given queries and keys from the backbone feature $\mI$, the keys are convolved by the queries to generate segmentation masks. The masks go through NMS to form the object mask outputs.
}
\label{fig:free_mask}
\vspace{-0.5em}
\end{figure}

Free Mask generates object masks from unlabeled images. 
As shown in Figure~\ref{fig:free_mask}, given an input image, dense feature maps $\mI\in \mathbb{ R}^{H\times W\times E}$ are extracted by a backbone model trained via self-supervision, \eg, ResNet~\cite{resnet} or any other convolutional neural network. 
This pre-trained model can be from supervised or unsupervised pre-training, as discussed below.
We first construct queries $\mQ$ and keys $\mK$ from the features $\mI$, which work together to generate the coarse segmentation masks.
We bilinearly downsample $\mI$ to form the queries $\mQ\in \mathbb{ R}^{H'\times W'\times E}$, where $H'$  and $W'$  denote the downsampled spatial size.
$\mI$ itself is used as the set of keys $\mK$.
For each query in $\mQ$, 
we compute its cosine similarity with every key in $\mK$, thus %
obtaining 
the score maps $\mS\in \mathbb{ R}^{H\times W \times N}$, where $N = H' \times W'$ is the total number of queries.
This operation can be written as:
\begin{equation}
\label{eq:freemask_sim}
\mS_{i,j,q} = {\tt sim}(\bm \mQ_q,\bm \mK_{i, j}),
\end{equation}
where $\bm \mQ_q\in \mathbb{ R}^{E}$ is the $q^\text{th}$ query, and $\mK_{i, j}\in \mathbb{ R}^{E}$ is the key at spatial location $(i, j)$.
${\tt sim}(\bm u,\bm v)$ denotes the cosine similarity, calculated by the dot product between $\ell_2$-normalized $\bm u$ and $\bm v$, \ie, ${\tt sim}(\bm u,\bm v) = \bm u^\top \bm v / \lVert\bm u\rVert \lVert\bm v\rVert$.
The process can also be viewed as a convolution where the $\ell_2$  normalized queries ${\mQ}'$  and  keys ${\mK}'$  are respectively the convolutional kernels and the features to be convolved together.
Each of the normalized queries is treated as a $1\times 1$ convolutional kernel.
Thus the operation can also be written as:
\begin{equation}
\begin{aligned}
\mS = {\mQ}' \circledast {\mK}'.
\end{aligned}
\end{equation}
The score maps are then normalized as soft masks by shifting the scores to the range $[0, 1]$. 
We compute the `{$\tt maskness$}' score defined further below for each of the $N$  soft masks, which serves as a confidence score of each extracted mask.
The soft masks are converted to binary masks using a threshold $\tau$.
We then sort the binary masks by their maskness scores and remove the redundant masks via mask non-maximum-suppression (NMS).
The %
overall process can be formulated as:
\begin{equation}
\label{eq:freemask_pipeline}
\begin{aligned}
\mM = \tt{NMS}\bigl( Maskness( Norm({\mQ}' \circledast {\mK}')   )    \bigr),
\end{aligned}
\end{equation}
where $\mM$ denotes the object masks that Free Mask outputs.

\myparagraph{Self-supervised pre-training.}
Free Mask uses a pre-trained backbone via self-supervision as the starting point.
We propose to leverage the self-supervised model pre-trained with dense correspondence.
Specifically, we find that dense contrastive learning~\cite{wang2020DenseCL} achieves considerably better results with our Free Mask approach, compared to the conventional self-supervised learning by global image-level contrasting. 
This can be attributed to the similar objective of Free Mask and dense contrastive learning.
Here we briefly introduce how the dense contrastive learning is performed.
It optimizes a pairwise (dis)similarity loss at the level of local features between two views of the input image.
A local feature vector, \ie, a query vector, should be similar to the corresponding positive key  in the other view while being dissimilar to other negative keys.
Observe that this is also aligned with Equation~\eqref{eq:freemask_sim} where the cosine similarity between a query and the keys is evaluated.
This also explains why Free Mask extracts reasonable masks.
We believe that there could be even better pre-training methods for Free Mask, \eg, those which tackle how to learn fine-grained representations at higher resolutions to generate better masks. We leave this for future research.

\myparagraph{Pyramid queries.}
When constructing the queries $\mQ$ from~$\mI$, we design a pyramid queries method to generate masks for instances at different scales. 
Specifically, we set a list of scale factors, \eg, $[1.0, 0.5, 0.25]$, when downsampling $\mI$, thus leading to a list of $\mQ$ at different scales from large to small.
All pyramid queries are flattened and concatenated together as the final $\mQ$. 

\myparagraph{Maskness score.} 
A scoring function is required for evaluating the quality of each generated coarse mask, which cannot be learned from annotations.
We use the non-parametric maskness method~\cite{wang2020solo}, \ie, ${\tt  maskness} = \frac{1}{N_f} \sum_{i}^{N_f} {\mathbf{p}_{i}}$, to obtain the confidence score of an extracted mask.
Here $N_f$ denotes the number of foreground pixels of the soft mask $\mathbf{p}$, \ie, the pixels that have values greater than threshold $\tau$. 
Intuitively, this score weighs more heavily on masks that have high 
 confidence on foreground pixels and down weights masks with uncertain foreground pixels.

\myparagraph{Unified with SOLO.}
We %
can see 
that the pipeline in Equation~\eqref{eq:freemask_pipeline} is unified with that of SOLO, as introduced in the above background section.
They both go through FCN, dynamic convolution, normalization and NMS operations to generate object masks.
However,  
the two are proposed to solve different problems. 
The latter aims to learn instance segmentation with rich annotated data, while the former is for segmenting objects in unlabeled images.
This provides a unifying perspective on segmenting objects in images.

\subsection{Self-Supervised SOLO}

We aim to train the SOLO-based instance segmenter using the segmentation masks and semantic embeddings, \ie, feature embeddings with high-level semantics, from Free Mask.
We separately introduce the methods for learning with coarse masks, self-training, and the semantic representation learning.

\def\max{{\rm max}}
\def\avg{{\rm avg}}
\myparagraph{Learning with coarse masks.}
In SOLO, the Dice loss~\cite{vnet} is used to supervise the predicted masks with their ground truth labels.
However, this is not ideally suited for our case of learning with noisy masks.
As the masks are coarse, directly using them as ground-truth masks can lead to unsatisfactory results.
We propose to use the coarse masks as a type of weak annotation and perform weakly supervised instance segmentation with them.

Inspired by the latest weakly-supervised method of BoxInst \cite{tian2020boxinst}, we project the predicted masks and the coarse masks on to the $x$-axis and the $y$-axis via a \texttt{max} operation along each axis.
The model is supervised to minimize the discrepancy between the projections of predicted masks and the coarse masks.
The loss term can be defined as:
\begin{equation}
\begin{aligned}
\mathcal{L}_{max\_proj} &=   \mathcal{L}(\max_x(\vm), \max_x(\vm^{*})) \\
 &+  \mathcal{L}(\max_y(\vm), \max_y(\vm^{*})),
\end{aligned}
\end{equation}
where $\mathcal{L}(\cdot, \cdot)$ is the Dice loss, $\vm$ and $\vm^{*}$ are the predicted mask and the coarse mask.
 $\max_x$ and $\max_y$  denote the \texttt{max} operations along each axis.

We further propose to project the predicted and coarse masks onto the $x$ and $y$ axes via an \texttt{average} operation along each axis.
The motivation is that the \texttt{max} operation may emphasize outlier segmentations in coarse masks, while the \texttt{average} operation de-emphasize the outliers.
In addition, \texttt{average} operation preserves solid shape of the object mask, which can benefit the training.
The loss term can be written as:
\begin{equation}
\begin{aligned}
\mathcal{L}_{avg\_proj} &=   \mathcal{L}(\avg_x(\vm), \avg_x(\vm^{*}))\\
 &+  \mathcal{L}(\avg_y(\vm), \avg_y(\vm^{*})) ,
\end{aligned}
\end{equation}
where  $\avg_x$ and $\avg_y$  denote the \texttt{average} operation along each axis. 
We also employ a pairwise affinity loss $\mathcal{L}_{pairwise}$~\cite{tian2020boxinst} to leverage the prior that the proximal pixels are likely to be in the same class, \ie, foreground or background, if they have similar colors in the raw image.

Overall, the total loss for mask prediction can be formulated as:
\begin{equation}
\label{eq:loss_mask}
\begin{aligned}
\mathcal{L}_{mask} =  \alpha\mathcal{L}_{avg\_proj} + \mathcal{L}_{max\_proj} + \mathcal{L}_{pairwise},
\end{aligned}
\end{equation}
where $\alpha$ acts as the weight to balance the various loss terms.

\myparagraph{Self-training.} 
With our carefully-designed loss function, we are able to train a SOLO-based instance segmenter with the free and noisy coarse masks. 
As shown in Figure~\ref{fig:free_solo_pipeline}, the object masks predicted by the instance segmenter are considerably better than the original coarse masks from Free Mask, which is also validated by the boosted accuracy in Table~\ref{tab:ablation_selftrain}.
As such, we propose to perform self-training with the initially trained instance segmenter to further improve accuracy.
We input unlabeled images into the instance segmenter and collect their predicted object masks.
The low-confidence predictions are removed and the remaining ones are treated as a new set of coarse masks.
We again train an instance segmenter with the unlabeled images and the new masks, using the loss function in Equation~\eqref{eq:loss_mask}. 
Performing self-training once already brings clear improvements and more iterations do not provide additional gains.

\myparagraph{Semantic representation learning.}
General instance segmentation requires not only localizing objects at the pixel level, but also recognizing
their semantic categories. 
In SOLO, the category branch predicts the semantic categories (including background) for each of the objects.
In our case without annotations, we propose to decouple the category branch to perform two sub-tasks: foreground/background binary classification, and semantic embedding learning.
The former task is trained with the conventional Focal loss~\cite{focalloss}, termed $\mathcal{L}_{focal}$.
For the latter task, we propose a simple approach for learning object-level semantic representations.
From Free Mask (introduced in Section~\ref{sub:freemask}), in addition to the segmentation masks, we can also directly obtain the semantic embedding of the discovered objects.
As shown in Figure~\ref{fig:free_mask}, each mask is associated with a query feature vector $\bm \mQ_q\in \mathbb{ R}^{E}$.
When training the instance segmenter, we add a branch in parallel to the last layer of the original category branch, which consists of a single convolution layer to predict the semantic embedding of each object.
Given the predicted and extracted  embeddings $\vq$ and $\vq^{*}$, we train 
the model 
by minimizing their negative cosine similarity:
\begin{equation}
\label{eq:sememb_loss}
\begin{aligned}
L_{sem} = 1 - \frac{\mathbf{\vq}}{\lVert \mathbf{\vq} \rVert_2}  \cdot \frac{\mathbf{\vq^{*}}}{\lVert \mathbf{\vq^{*}} \rVert_2}.
\end{aligned}
\end{equation}
The total loss for the category branch can be written as:
\begin{equation}
\label{eq:loss_cate}
\begin{aligned}
\mathcal{L}_{cate} = \mathcal{L}_{focal} +  \beta\mathcal{L}_{sem},
\end{aligned}
\end{equation}
where $\beta$ acts as the weight to balance the two terms.
Overall, we train the instance segmenter with a combination of $\mathcal{L}_{mask}$ and $\mathcal{L}_{cate}$, corresponding to the losses for the mask branch and category branch, respectively.

\section{Experiments}
\label{sec:exp}
\vspace{-0.5em}

\subsection{Experimental Settings}

\myparagraph{Technical details.}
For Free Mask, the shorter side of the input image is set to 800 pixels.
Threshold $\tau$ is set to $0.5$.
DenseCL~\cite{wang2020DenseCL} with a pre-trained ResNet-50~\cite{resnet} architecture is adopted as the backbone unless specified.
Matrix NMS~\cite{wang2020solo} is used for mask NMS.
After NMS, we filter out the low-quality masks with a maskness threshold of 0.7.
When training the SOLO model, we initialize the backbone with the pre-trained model used in Free Mask.
We set the $\alpha$ and $\beta$ parameters to $0.1$ and $4.0$, respectively.
We employ the simple copy-paste strategy~\cite{copypaste} for data augmentation.
During self-training, we set the confidence threshold for removing the low-confidence predictions to $0.3$.

\myparagraph{Datasets.}
For \Ours, we use the images in COCO \texttt{train2017} and COCO \texttt{unlabeled2017}~\cite{coco} as the set of unlabeled images, containing a total of $\app$241k images.
These unlabeled images are input to Free Mask and are used to train the instance segmenter.
The self-supervised backbone in Free Mask is pre-trained on ImageNet with \app1.28 million unlabeled images.
We further employ COCO \texttt{val2017}, UVO \texttt{val}~\cite{uvo}, and PASCAL VOC \texttt{trainval07}~\cite{voc} datasets for evaluation.

\myparagraph{Evaluation protocol.}
We evaluate self-supervised instance segmentation with the standard COCO protocol.
We report class-agnostic COCO mask average precision (AP) and average recall (AR) on $5$k \texttt{val2017} split, which is averaged over 10 intersection-over-union (IoU) thresholds evenly-spaced between $0.5$ and $0.95$. 
AP considers recall and precision simultaneously, which computes the average precision value for recall values over 0 to 1.
AR allows redundant or random detection results, as it computes the maximum recall given a fixed number of detections per image. 

To compare with unsupervised object detection methods, we convert the masks to boxes and report the box AP on both the COCO  \texttt{val2017}, COCO \texttt{20k}, and VOC \texttt{trainval07}.
We further evaluate the pre-trained model by fine-tuning with annotations.
Specifically, we fine-tune the instance segmenter on  COCO \texttt{train2017}  and evaluate on  COCO \texttt{val2017}.
We provide two settings, \ie, limited fully annotated images, and limited segmentation masks (see Appendix~\ref{app:ft}).
Mask AP averaged across all 10 IoU thresholds and all 80 categories is reported.

\begin{table}[bt]
\small
\tablestyle{5pt}{1.1}
\centering
\begin{tabular}{l|ccc|ccc}
\toprule
 method &  AP$_\text{50}$ & AP$_\text{75}$ & AP & AR$_\text{1}$ & AR$_\text{10}$ & AR$_\text{100}$ \\ 
\midrule
\emph{w/ anns:} &&&\\
MCG~\cite{mcg}  & 4.6 &	0.8 & 1.6 &	1.9 &	7.4 & 18.2 \\
COB~\cite{cob}  & 8.8 &	1.9 & 3.3 &	2.9 &	10.1 & 22.7 \\
\midrule
\emph{w/o anns:} &&&\\
\textbf{\Ours}      & 9.8  & 2.9 & 4.0 &  4.1  & 10.5 & 12.7\\
\bottomrule
\end{tabular}
\vspace{-0.5em}
\caption{\textbf{Class-agnostic instance segmentation}  on MS COCO \texttt{val2017}. 
Both MCG and COB require annotations more or less.
}
\label{tab:insseg_val}
\vspace{-1.0em}
\end{table}

\begin{table}[bt]
\small
\tablestyle{5pt}{1.1}
\centering
\begin{tabular}{l|ccccccc}
\toprule
 method  &  AP$_\text{50}$ & AP$_\text{75}$ & AP \\ 
\midrule
\emph{w/ full anns:} &&&\\
SOLOv2 w/ COCO & 38.0 & 20.9 & 21.4 \\
Mask R-CNN w/ COCO &  31.0 & 14.2 & 15.9\\
SOLOv2 w/ LVIS & 14.8 & 5.9 & 7.1 \\
Mask R-CNN w/ LVIS & 18.1 & 4.1 &  6.8\\
\midrule
\emph{w/o anns:} &&&\\
\textbf{\Ours}   & 12.7  & 3.0 & 4.8  \\
\bottomrule
\end{tabular}
\vspace{-0.5em}
\caption{\textbf{Class-agnostic instance segmentation} on UVO \texttt{val} split.
Results of Mask R-CNN are from the paper of UVO~\cite{uvo}.
}
\label{tab:insseg_uvo}
\vspace{-1.5em}
\end{table}

\subsection{Main Results}
\label{subsec:main_results}

\myparagraph{Self-supervised instance segmentation.}
For evaluating the self-supervised instance segmenter, we first provide qualitative results to show how \Ours performs at the task of class-agnostic instance segmentation. 
As shown in Figure~\ref{fig:vis_first}, without any annotations, \Ours is able to segment object instances of many different categories. 
To provide a quantitative comparison with previous methods, we report the results of unsupervised class-agnostic instance segmentation in Table~\ref{tab:insseg_val} and Table~\ref{tab:insseg_uvo}.
As there is no reported result for this new problem, we evaluate a few popular segmentation proposal methods on this benchmark.
Among the compared methods, MCG~\cite{mcg} uses the annotated BSDS500 dataset~\cite{MartinFTM01} for training a boundary detector, and COB~\cite{cob} trains its hierarchies and combinatorial grouping on PASCAL Context dataset~\cite{MottaghiCLCLFUY14}.
By contrast, our \Ours method achieves better results without any annotations. 
We further compare against the supervised methods trained with full annotations.
It is worth noting that \Ours even performs closely to the fully supervised Mask R-CNN~\cite{he2017mask} trained on
the 
LVIS dataset~\cite{lvis2019}, \eg, 4.8\% vs 6.8\% AP on the 
UVO dataset.

\myparagraph{Self-supervised object detection.}
By converting the masks into boxes, our self-supervised instance segmenter naturally serves as a self-supervised object detector as well.
We report the results of class-agnostic object detection on COCO \texttt{val2017} benchmark in Table~\ref{tab:det_val}.
Our method shows significantly superior performance.
To compare with existing object discovery methods, we also evaluate \Ours on VOC \texttt{trainval07} and COCO \texttt{20k} for multi-object discovery.
As shown in Table~\ref{tab:det_discovery}, our method largely outperforms the state-of-the-art object discovery methods, including a concurrent work~\cite{lost}.
Its relative improvements are up to 100\% on the COCO dataset.

\begin{table}[t]
\small
\tablestyle{5pt}{1.1}
\centering
\begin{tabular}{l|ccc|ccc}
\toprule
method  &  AP$_\text{50}$ & AP$_\text{75}$  & AP & AR$_\text{1}$ & AR$_\text{10}$ & AR$_\text{100}$ \\ 
\midrule
UP-DETR~\cite{updetr} & 0.0  & 0.0 & 0.0 & 0.0 & 0.0 & 0.4 \\
Selective Search~\cite{selectivesearch}  & 0.5 & 0.1 & 0.2 & 0.2 & 1.5 & 10.9\\
DETReg~\cite{bar2021detreg}   & 3.1 & 0.6 & 1.0 & 0.6 & 3.6 & 12.7 \\
\textbf{\Ours}  &  12.2 & 4.2   & 5.5 & 4.6 & 11.4 & 15.3 \\
\bottomrule
\end{tabular}
\vspace{-0.56em}
\caption{\textbf{Unsupervised class-agnostic object detection }on MS COCO \texttt{val2017}. 
Compared results are directly from DETReg.}
\label{tab:det_val}
\vspace{-1.0em}
\end{table}

\begin{table}[t]
\small
\tablestyle{5pt}{1.1}
\centering
\begin{tabular}{l|ccc|ccc}
\toprule
 \multirow{2}{*}{method}  & \multicolumn{3}{c}{VOC} & \multicolumn{3}{c}{COCO} \\
 \cline{2-7} 
  &  AP$_\text{50}$ & AP$_\text{75}$  & AP  &  AP$_\text{50}$ & AP$_\text{75}$ &  AP \\ 
\midrule
Kim et al.~\cite{KimT09}  & 9.5 & - & 2.5 & 3.9 & - &  1.0  \\
DDT+~\cite{wei2019unsupervised}  & 8.7 & -  &  3.0  & 2.4 & - & 0.7 \\
rOSD~\cite{vo2020toward}  & 13.1 & - &  4.3  & 5.2  & - & 1.6 \\
LOD~\cite{lod} & 13.9 & -  & 4.5   & 6.6 & - & 2.0 \\
LOST*~\cite{lost} & 19.8 & -  & 6.7   & 7.9 & - & 2.5 \\
\textbf{\Ours}  & 24.5  & 7.2  &  10.2  &  12.4 & 4.4 & 5.6 \\
\bottomrule
\end{tabular}
\vspace{-0.5em}
\caption{\textbf{Multi-object discovery} on PASCAL VOC \texttt{trainval07} and MS COCO \texttt{20k}. 
LOST* is a concurrent work.
}
\label{tab:det_discovery}
\vspace{-0.5em}
\end{table}

\begin{table}[t]
\small
\tablestyle{5pt}{1.1}
\centering
\begin{tabular}{cl|cccccc}
\toprule
\multicolumn{2}{c|}{pre-train} & AP &  AP$_\text{50}$ & AP$_\text{75}$ &AP$_{S}$ & AP$_{M}$ & AP$_{L}$ \\ 
\midrule
\multirow{4}{*}{\rotatebox{90}{\emph{5\% images}}} &
sup. & 18.0 & 32.2 & 17.6 & 5.5 & 18.9 & 27.8\\
& MoCo-v2~\cite{mocov2} &19.0 & 32.7 & 19.2 & 5.4 & 19.9 & 28.9\\
& DenseCL~\cite{wang2020DenseCL}  & 20.0 & 33.7 & 20.5 & 5.5 & 21.5 & 30.1  \\
& \textbf{\Ours} & 22.0 & 36.0 & 22.9 & 6.5 & 23.2 & 33.8 \\
\midrule
\multirow{4}{*}{\rotatebox{90}{\emph{10\% images}}}  &
sup. & 22.3 & 38.0 & 22.9 & 6.3  & 24.0 & 34.8 \\
& MoCo-v2~\cite{mocov2} & 23.2 & 39.0 & 23.9 & 6.7 & 24.6 & 36.2 \\
& DenseCL~\cite{wang2020DenseCL}  & 23.7 &  39.3 & 24.5 & 7.3 & 25.2 & 37.1  \\
& \textbf{\Ours}  &  25.6 &  41.6 &  26.7 & 8.3 & 27.5 & 40.3 \\
\bottomrule
\end{tabular}
\vspace{-0.5em}
\caption{
\textbf{Supervised instance segmentation} with limited fully annotated images. 
}
\label{tab:ft_image}
\vspace{-0.5em}
\end{table}

\begin{table}[t]
\small
\tablestyle{5pt}{1.1}
\centering
\begin{tabular}{cl|cccccc}
\toprule
\multicolumn{2}{c|}{pre-train} & AP &  AP$_\text{50}$ & AP$_\text{75}$ &AP$_{S}$ & AP$_{M}$ & AP$_{L}$ \\ 
\midrule
\multirow{4}{*}{\rotatebox{90}{\emph{5\% masks}}} &
sup. & 17.8 &  36.1 & 15.9 & 6.3 & 19.5 & 27.4 \\
& MoCo-v2~\cite{mocov2} & 17.2 &  34.9 & 14.9 & 5.8 & 19.0 & 26.2\\
& DenseCL~\cite{wang2020DenseCL}  & 20.1 & 39.0 & 18.3 & 7.6 & 21.4 & 31.2 \\
& \textbf{\Ours} &29.9 & 50.5 & 30.5 & 10.7 & 32.5 & 46.7 \\
\midrule
\multirow{4}{*}{\rotatebox{90}{\emph{10\% masks}}} &
sup. & 25.4 & 45.6 & 25.1 & 8.8 & 26.9 & 40.7\\
& MoCo-v2~\cite{mocov2} & 25.6 & 45.1 & 25.5 & 8.7 & 27.2 & 40.4\\
& DenseCL~\cite{wang2020DenseCL}  & 26.1 & 45.2 & 26.3 & 9.1 & 28.0 & 40.8\\
& \textbf{\Ours}  &  31.1 &  51.4 & 32.0 & 11.2 & 34.1 & 48.4\\
\bottomrule
\end{tabular}
\vspace{-0.5em}
\caption{\textbf{Supervised instance segmentation}  with limited segmentation masks. 
}
\label{tab:ft_mask}
\vspace{-1.0em}
\end{table}

\myparagraph{Supervised fine-tuning.}
In addition to evaluating the self-supervised instance segmenter directly,
we also evaluate the performance of our approach in a supervised setting by fine-tuning the self-supervised instance segmenter with annotations. As shown in Table~\ref{tab:ft_image}, \Ours pre-training outperforms ImageNet supervised pre-training by 4.0\% AP when using 5\% COCO training images.
The gains over the state-of-the-art self-supervised pre-training methods are also clear, \eg, 2.0\% AP better than DenseCL~\cite{wang2020DenseCL}.

To further compare the pre-training methods with different amount of mask annotations, in Table~\ref{tab:ft_mask}, we conduct fine-tuning experiments with only limited masks available.
When fine-tuning with 5\% masks, 
\Ours achieves significant gains of 9.8\% AP over supervised pre-training.
These fine-tuning experiments demonstrate that \Ours serves as a strong instance segmentation pre-training method, outperforming both the supervised and state-of-the-art self-supervised pre-training methods.

\iftrue
\begin{table*}[t]
\vspace{-.2em}
\centering
\subfloat[
\footnotesize
\textbf{Different pre-training methods} with Free Mask.
DenseCL works the best.
\label{tab:ablation_freemask_pretrain}
]{
\centering
\begin{minipage}{0.3\linewidth}{\begin{center}
\tablestyle{4pt}{1.05}
\begin{tabular}{l|cccc}
\toprule
 pre-train & AR &AR$_{S}$ & AR$_{M}$ & AR$_{L}$ \\ 
\midrule
sup. & 7.8 & 0.1 & 11.3	& 16.4 \\
SimCLR~\cite{simclr}  & 6.1	& 1.0 & 12.1 & 6.7 \\
MoCo-v2~\cite{mocov2}  & 4.7 & 1.6 & 8.1 & 5.4 \\
DINO~\cite{dino}  & 3.2 & 2.8 & 5.2 & 0.9 \\
EsViT~\cite{li2021esvit} & 6.3	 & 0.0 & 6.0 & 17.8 \\
DenseCL~\cite{wang2020DenseCL}  & 11.5 & 0.1 & 6.0 & 39.5  \\
\bottomrule
\end{tabular}
\end{center}}\end{minipage}
}
\hspace{1.5em}
\subfloat[
\footnotesize
\textbf{Pyramid queries} in Free Mask.
Pyramid queries improve over single scale queries.
\label{tab:ablation_freemask_pyramid}
]{
\begin{minipage}{0.3\linewidth}{\begin{center}
\tablestyle{4pt}{1.05}
\begin{tabular}{l|cccc}
\toprule
 scale & AR &AR$_{S}$ & AR$_{M}$ & AR$_{L}$ \\ 
\midrule
0.25 & 10.1 &0.0 & 1.9 & 39.5\\
1.0 & 11.3 & 0.1 & 6.0 & 38.6 \\
pyramid & 11.5 & 0.1 & 6.0 & 39.5  \\
\bottomrule
\end{tabular}
\vspace{3.75em}
\end{center}}\end{minipage}
}
\centering
\hspace{1.5em}
\subfloat[
\footnotesize
\textbf{Self-training iterations.}
`-1' refers to coarse masks.
`0' means learning without self-training.
\label{tab:ablation_selftrain}
]{
\begin{minipage}{0.3\linewidth}{\begin{center}
\tablestyle{4pt}{1.05}
\begin{tabular}{c|ccccccc}
\toprule
 iters &  AP$_\text{50}$ & AP$_\text{75}$ & AP \\ 
\midrule
-1 & 2.3 & 0.2 & 0.7  \\
0 &  7.9 & 2.5 & 3.3 \\
1  & 8.3 & 2.8 & 3.7 \\
2 &  7.7 &	2.9 & 3.5\\
\bottomrule
\end{tabular}
\vspace{2.6em}
\end{center}}\end{minipage}
}
\\
\vspace{0.1em}

\subfloat[
\footnotesize
\textbf{Full vs.\ weak supervision.}
Weakly-supervised design is effective.
\label{tab:ablation_fullvsweak}
]{
\centering
\begin{minipage}{0.3\linewidth}{\begin{center}
\tablestyle{4pt}{1.05}
\begin{tabular}{l|ccccccc}
\toprule
 mask loss  &  AP$_\text{50}$ & AP$_\text{75}$ & AP  \\ 
\midrule
full  &  6.2& 1.6 & 2.4\\
weak &  7.9 & 2.5 & 3.3 \\
\bottomrule
\end{tabular}
\end{center}}\end{minipage}
}
\hspace{1.5em}
\subfloat[
\footnotesize
\textbf{Mask loss terms.}
Each loss component contributes to the final results.
\label{tab:ablation_lossterms}
]{
\centering
\begin{minipage}{0.3\linewidth}{\begin{center}
\tablestyle{4pt}{1.05}
\begin{tabular}{l|ccccccc}
\toprule
 mask loss  &  AP$_\text{50}$ & AP$_\text{75}$ & AP  \\ 
\midrule
combination &  7.9 & 2.5 & 3.3 \\
-  w/o $\mathcal{L}_{avg\_proj}$  &  3.8 & 1.6 & 2.0 \\
- w/o $\mathcal{L}_{max\_proj}$  &  	7.1 &	1.6 & 2.6\\
- w/o $\mathcal{L}_{pairwise}$  &  	6.1 &	0.9 & 2.1\\
\bottomrule
\end{tabular}
\end{center}}\end{minipage}
}
\hspace{1.5em}
\subfloat[
\footnotesize
\textbf{Semantic embedding. }
Semantic embedding learning improves the fine-tuning results.
\label{tab:ablation_emb}
]{
\begin{minipage}{0.3\linewidth}{\begin{center}
\tablestyle{4pt}{1.05}
\begin{tabular}{l|ccc}
\toprule
 $\mathcal{L}_{sem}?$ & AP &  AP$_\text{50}$ & AP$_\text{75}$  \\ 
\midrule
 & 24.9 & 40.5 & 26.1  \\
\checkmark & 25.6 & 41.6 & 26.7 \\
\bottomrule
\end{tabular}
\end{center}}\end{minipage}
}

\vspace{-.3em}
\caption{\textbf{\Ours ablation experiments.} All the experiments are with a ResNet-50 backbone. We report class-agnostic instance segmentation results (a-e) and supervised fine-tuning results (f) on
the 
COCO \texttt{val2017} split.}
\label{tab:ablations} 
\vspace{-.5em}
\end{table*}
\fi

\subsection{Ablation Study}
\label{subsec:ablation}
\vspace{-0.5em}

We conduct %
ablation experiments to show how
 each component contributes
 to \Ours.
The ablation studies are performed on the COCO \texttt{val2017} split.

\begin{figure}[t!]
\centering
\includegraphics[width=0.75\linewidth]{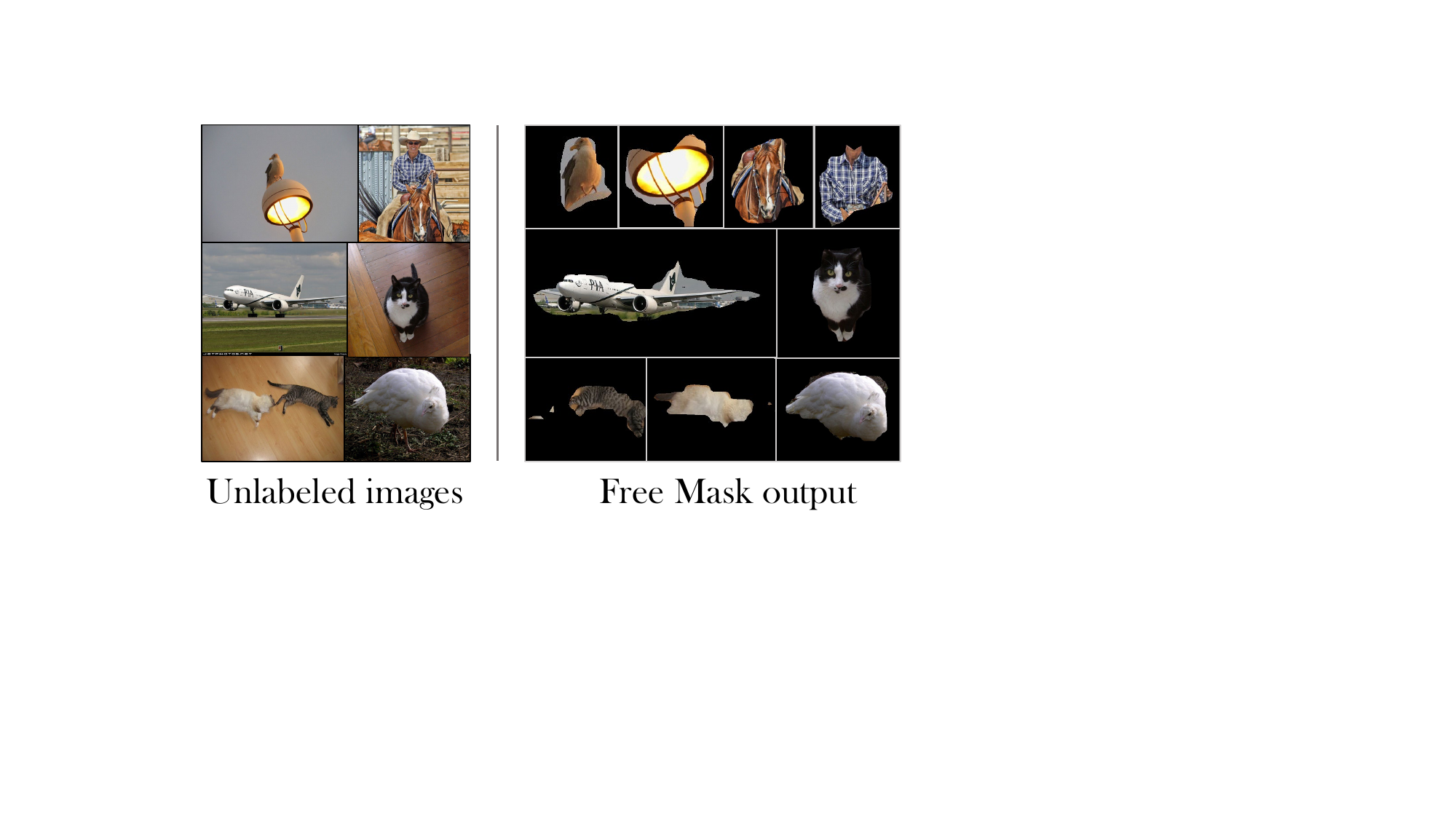}
\vspace{-0.5em}
\caption{
Qualitative results of the Free Mask. 
Free Mask extracts coarse masks of the common objects in unlabeled images.
}
\label{fig:freemask_vis}
\vspace{-1.2em}
\end{figure}

\myparagraph{Free Mask with different pre-trained backbones.}
In Table~\ref{tab:ablation_freemask_pretrain}, we show how Free Mask performs with different pre-trained backbones.
The  conventional self-supervised learning methods that contrast the global representations of image pairs, \eg, SimCLR and MoCo-v2, show worse results compared to supervised ImageNet pre-training. 
The self-supervised learning methods that consider dense correspondence, \eg, EsViT and DenesCL, yield better results than those that do not.
DenseCL shows the best results compared to both supervised and other self-supervised methods.
This aligns with our hypothesis  in Section~\ref{sub:freemask} that DenseCL's objective is consistent with Free Mask's.
We provide some visualizations of Free Mask in Figure~\ref{fig:freemask_vis}.

\myparagraph{Pyramid queries.}
We compare different scales of the queries $\mQ$ used in Free Mask in Table~\ref{tab:ablation_freemask_pyramid}.
A smaller scale is better for large objects but worse for medium and small objects.
A large scale is just the opposite. 
Pyramid queries with scales $[1.0, 0.5, 0.25]$ yield the best results.

\begin{figure}[t!]
\centering
\includegraphics[width=0.76\linewidth]{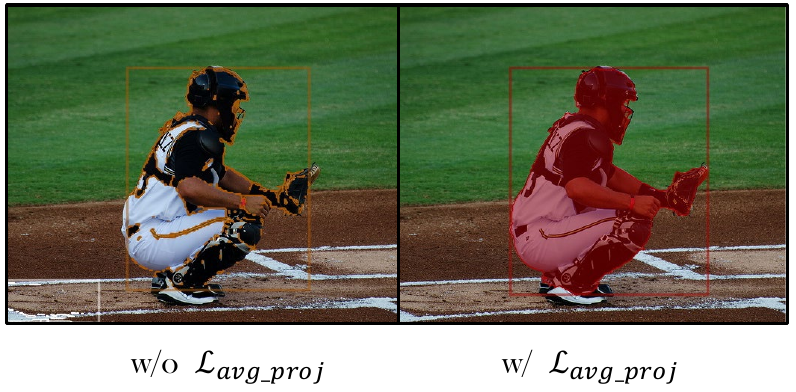}
\vspace{-0.5em}
\caption{
Qualitative comparison of with and without $\mathcal{L}_{avg\_proj}$ when learning from coarse masks.
The model trained without $\mathcal{L}_{avg\_proj}$ tends to only segment the contours when trained longer. %
}
\label{fig:avgproj_vis}
\vspace{-1.0em}
\end{figure}

\myparagraph{Loss functions.}
In Table~\ref{tab:ablation_fullvsweak}, we compare our weakly-supervised design against the full mask supervision, \ie, the original Dice loss used in SOLO computed with the full masks.
Directly using the coarse masks to provide full supervision to the instance segmenter leads to unsatisfactory results.
Our weakly-supervised loss outperforms the original full mask loss by a large margin.
In Table~\ref{tab:ablation_lossterms}, we study the mask loss terms in Equation~\eqref{eq:loss_mask}.
The performance drops sharply when learning without $\mathcal{L}_{avg\_proj}$, \ie, with only the projection loss from \texttt{max} operation  and pairwise loss as in~\cite{tian2020boxinst}.
The model even collapses to only segmenting the contours when trained longer (Figure~\ref{fig:avgproj_vis}). 
Our method tackles this problem by leveraging the projection from \texttt{average} operation, which not only preserves the shape but is also less sensitive to outlier pixels.

\myparagraph{Self-training.}
Our method performs self-training by selecting high-confidence predictions of the self-supervised instance segmenter and training the instance segmenter again with them.
We compare the results of performing different iterations of self-training in Table~\ref{tab:ablation_selftrain}.
`$-1$' refers to the initial coarse masks.
Zero iteration refers to learning from the coarse masks without self-training.
We show that performing self-training once already brings clear improvements, but additional iterations do not provide additional gains.

\myparagraph{Semantic embedding.}
To validate the effectiveness of the semantic embedding learning, in Table~\ref{tab:ablation_emb} we compare the models trained with or without the semantic embedding loss defined in Equation~\eqref{eq:sememb_loss}.
The models are fine-tuned with 10\% of fully annotated COCO images.
It shows that the semantic embedding loss yields clear improvements when fine-tuning instance segmentation with annotations.

\begin{figure}[t!]
\centering
\includegraphics[width=0.75\linewidth]{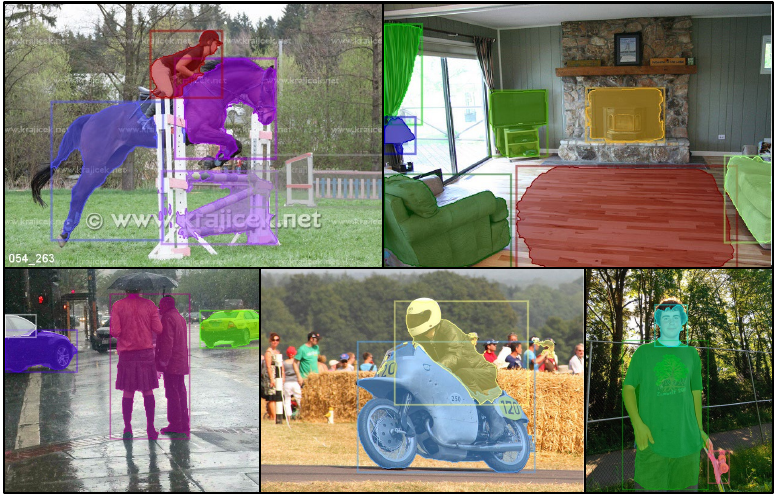}
\vspace{-0.5em}
\caption{
Failure cases of \Ours. Our method could fail to localize objects that are truncated, crowded or small. 
}
\label{fig:vis_failure}
\vspace{-1.0em}
\end{figure}

\vspace{-1.0em}
\section{Discussion and Conclusion}
\vspace{-0.6em}
In this work, we have developed a simple and effective self-supervised instance segmentation framework \Ours. \Ours enables learning to segment objects without any annotations, neither pixel-level nor image-level labels.
We hope that its novel design elements provide insights for future works on unsupervised visual learning, \eg, unsupervised panoptic segmentation, and beyond.

\myparagraph{Limitations.}
Without category labels, our self-supervised instance segmenter cannot predict the categories of the detected objects, but generate class-agnostic object masks.
There is still a large gap between our self-supervised model and the supervised one trained with rich annotations.
Our method could fail in some scenarios (Figure~\ref{fig:vis_failure}).
We believe there is plenty of room to improve based on our method.

\myparagraph{Broader impacts.}
This work shows that one can learn a class-agnostic instance segmenter without any annotations.
In the future, there is a chance for self-supervised  segmenter to reach or even outperform the supervised model trained with manual annotations, which may eliminate the need for annotating  masks or boxes for common objects.
We expect that the proposed technique can be used to largely reduce data annotation effort for a few instance-level recognition tasks in computer vision.

{\small
\bibliographystyle{unsrt}
\bibliography{egbib}

\begin{thebibliography}{10}

\bibitem{fcis}
Yi~Li, Haozhi Qi, Jifeng Dai, Xiangyang Ji, and Yichen Wei.
\newblock Fully convolutional instance-aware semantic segmentation.
\newblock In {\em IEEE Conf. Comput. Vis. Pattern Recog.}, 2017.

\bibitem{he2017mask}
Kaiming He, Georgia Gkioxari, Piotr Doll{\'a}r, and Ross Girshick.
\newblock {Mask R-CNN}.
\newblock In {\em Int. Conf. Comput. Vis.}, 2017.

\bibitem{de2017semantic}
Bert De~Brabandere, Davy Neven, and Luc Van~Gool.
\newblock Semantic instance segmentation with a discriminative loss function.
\newblock {\em arXiv:1708.02551}, 2017.

\bibitem{chen2019hybrid}
Kai Chen, Jiangmiao Pang, Jiaqi Wang, Yu~Xiong, Xiaoxiao Li, Shuyang Sun,
  Wansen Feng, Ziwei Liu, Jianping Shi, Wanli Ouyang, et~al.
\newblock Hybrid task cascade for instance segmentation.
\newblock In {\em IEEE Conf. Comput. Vis. Pattern Recog.}, 2019.

\bibitem{yolact}
Daniel Bolya, Chong Zhou, Fanyi Xiao, and Yong~Jae Lee.
\newblock Yolact: {Real-time} instance segmentation.
\newblock In {\em Int. Conf. Comput. Vis.}, 2019.

\bibitem{CondInst}
Zhi Tian, Chunhua Shen, and Hao Chen.
\newblock Conditional convolutions for instance segmentation.
\newblock In {\em Eur. Conf. Comput. Vis.}, 2020.

\bibitem{wang2021SOLO}
Xinlong Wang, Rufeng Zhang, Chunhua Shen, Tao Kong, and Lei Li.
\newblock {SOLO}: A simple framework for instance segmentation.
\newblock {\em IEEE Trans. Pattern Anal. Mach. Intell.}, 2021.

\bibitem{KhorevaBH0S17}
Anna Khoreva, Rodrigo Benenson, Jan~Hendrik Hosang, Matthias Hein, and Bernt
  Schiele.
\newblock Simple does it: Weakly supervised instance and semantic segmentation.
\newblock In {\em IEEE Conf. Comput. Vis. Pattern Recog.}, 2017.

\bibitem{hsu2019weakly}
Cheng-Chun Hsu, Kuang-Jui Hsu, Chung-Chi Tsai, Yen-Yu Lin, and Yung-Yu Chuang.
\newblock Weakly supervised instance segmentation using the bounding box
  tightness prior.
\newblock In {\em Adv. Neural Inform. Process. Syst.}, 2019.

\bibitem{liu2020leveraging}
Yun Liu, Yu-Huan Wu, Peisong Wen, Yujun Shi, Yu~Qiu, and Ming-Ming Cheng.
\newblock Leveraging instance-, image- and dataset-level information for weakly
  supervised instance segmentation.
\newblock {\em IEEE Trans. Pattern Anal. Mach. Intell.}, 2020.

\bibitem{tian2020boxinst}
Zhi Tian, Chunhua Shen, Xinlong Wang, and Hao Chen.
\newblock {BoxInst}: High-performance instance segmentation with box
  annotations.
\newblock In {\em IEEE Conf. Comput. Vis. Pattern Recog.}, 2021.

\bibitem{cheng2021pointly}
Bowen Cheng, Omkar Parkhi, and Alexander Kirillov.
\newblock Pointly-supervised instance segmentation.
\newblock {\em arXiv preprint arXiv:2104.06404}, 2021.

\bibitem{lan2021discobox}
Shiyi Lan, Zhiding Yu, Christopher Choy, Subhashree Radhakrishnan, Guilin Liu,
  Yuke Zhu, Larry~S Davis, and Anima Anandkumar.
\newblock Discobox: Weakly supervised instance segmentation and semantic
  correspondence from box supervision.
\newblock In {\em Int. Conf. Comput. Vis.}, 2021.

\bibitem{wang2020DenseCL}
Xinlong Wang, Rufeng Zhang, Chunhua Shen, Tao Kong, and Lei Li.
\newblock Dense contrastive learning for self-supervised visual pre-training.
\newblock In {\em IEEE Conf. Comput. Vis. Pattern Recog.}, 2021.

\bibitem{moco}
Kaiming He, Haoqi Fan, Yuxin Wu, Saining Xie, and Ross Girshick.
\newblock Momentum contrast for unsupervised visual representation learning.
\newblock In {\em IEEE Conf. Comput. Vis. Pattern Recog.}, 2020.

\bibitem{simclr}
Ting Chen, Simon Kornblith, Mohammad Norouzi, and Geoffrey Hinton.
\newblock A simple framework for contrastive learning of visual
  representations.
\newblock In {\em Int. Conf. Mach. Learn.}, 2020.

\bibitem{byol}
Jean{-}Bastien Grill, Florian Strub, Florent Altch{\'{e}}, Corentin Tallec,
  Pierre~H. Richemond, Elena Buchatskaya, Carl Doersch, Bernardo~{\'{A}}vila
  Pires, Zhaohan Guo, Mohammad~Gheshlaghi Azar, Bilal Piot, Koray Kavukcuoglu,
  R{\'{e}}mi Munos, and Michal Valko.
\newblock Bootstrap your own latent - {A} new approach to self-supervised
  learning.
\newblock In {\em Adv. Neural Inform. Process. Syst.}, 2020.

\bibitem{updetr}
Zhigang Dai, Bolun Cai, Yugeng Lin, and Junying Chen.
\newblock Up-detr: Unsupervised pre-training for object detection with
  transformers.
\newblock In {\em IEEE Conf. Comput. Vis. Pattern Recog.}, 2021.

\bibitem{detcon2021}
Olivier~J. H{\'{e}}naff, Skanda Koppula, Jean{-}Baptiste Alayrac, A{\"{a}}ron
  van~den Oord, Oriol Vinyals, and Jo{\~{a}}o Carreira.
\newblock Efficient visual pretraining with contrastive detection.
\newblock In {\em Int. Conf. Comput. Vis.}, 2021.

\bibitem{PinheiroABGC20}
Pedro~O. Pinheiro, Amjad Almahairi, Ryan~Y. Benmalek, Florian Golemo, and
  Aaron~C. Courville.
\newblock Unsupervised learning of dense visual representations.
\newblock In {\em Adv. Neural Inform. Process. Syst.}, 2020.

\bibitem{ChaitanyaEKK20}
Krishna Chaitanya, Ertunc Erdil, Neerav Karani, and Ender Konukoglu.
\newblock Contrastive learning of global and local features for medical image
  segmentation with limited annotations.
\newblock In {\em Adv. Neural Inform. Process. Syst.}, 2020.

\bibitem{panet}
Shu Liu, Lu~Qi, Haifang Qin, Jianping Shi, and Jiaya Jia.
\newblock Path aggregation network for instance segmentation.
\newblock In {\em IEEE Conf. Comput. Vis. Pattern Recog.}, 2018.

\bibitem{associativeembedding}
Alejandro Newell, Zhiao Huang, and Jia Deng.
\newblock Associative embedding: End-to-end learning for joint detection and
  grouping.
\newblock In {\em Adv. Neural Inform. Process. Syst.}, 2017.

\bibitem{SGN17}
Shu Liu, Jiaya Jia, Sanja Fidler, and Raquel Urtasun.
\newblock Sequential grouping networks for instance segmentation.
\newblock In {\em Int. Conf. Comput. Vis.}, 2017.

\bibitem{Gao_2019_ICCV}
Naiyu Gao, Yanhu Shan, Yupei Wang, Xin Zhao, Yinan Yu, Ming Yang, and Kaiqi
  Huang.
\newblock Ssap: Single-shot instance segmentation with affinity pyramid.
\newblock In {\em Int. Conf. Comput. Vis.}, 2019.

\bibitem{chen2020blendmask}
Hao Chen, Kunyang Sun, Zhi Tian, Chunhua Shen, Yongming Huang, and Youliang
  Yan.
\newblock {BlendMask}: Top-down meets bottom-up for instance segmentation.
\newblock In {\em IEEE Conf. Comput. Vis. Pattern Recog.}, 2020.

\bibitem{wang2020solo}
Xinlong Wang, Tao Kong, Chunhua Shen, Yuning Jiang, and Lei Li.
\newblock {SOLO}: Segmenting objects by locations.
\newblock In {\em Eur. Conf. Comput. Vis.}, 2020.

\bibitem{wang2020solov2}
Xinlong Wang, Rufeng Zhang, Tao Kong, Lei Li, and Chunhua Shen.
\newblock Solov2: Dynamic and fast instance segmentation.
\newblock {\em Adv. Neural Inform. Process. Syst.}, 2020.

\bibitem{Zhou2018PRM}
Yanzhao Zhou, Yi~Zhu, Qixiang Ye, Qiang Qiu, and Jianbin Jiao.
\newblock Weakly supervised instance segmentation using class peak response.
\newblock In {\em IEEE Conf. Comput. Vis. Pattern Recog.}, 2018.

\bibitem{zhang2016colorful}
Richard Zhang, Phillip Isola, and Alexei Efros.
\newblock Colorful image colorization.
\newblock In {\em Eur. Conf. Comput. Vis.}, 2016.

\bibitem{inpainting16}
Deepak Pathak, Philipp Krahenbuhl, Jeff Donahue, Trevor Darrell, and Alexei
  Efros.
\newblock Context encoders: Feature learning by inpainting.
\newblock In {\em IEEE Conf. Comput. Vis. Pattern Recog.}, 2016.

\bibitem{jigsaw}
Mehdi Noroozi and Paolo Favaro.
\newblock Unsupervised learning of visual representations by solving jigsaw
  puzzles.
\newblock In {\em Eur. Conf. Comput. Vis.}, 2016.

\bibitem{gidaris2018rotations}
Spyros Gidaris, Praveer Singh, and Nikos Komodakis.
\newblock Unsupervised representation learning by predicting image rotations.
\newblock In {\em Int. Conf. Learn. Represent.}, 2018.

\bibitem{wu2018unsupervised}
Zhirong Wu, Yuanjun Xiong, Stella Yu, and Dahua Lin.
\newblock Unsupervised feature learning via non-parametric instance
  discrimination.
\newblock In {\em IEEE Conf. Comput. Vis. Pattern Recog.}, 2018.

\bibitem{swav20}
Mathilde Caron, Ishan Misra, Julien Mairal, Priya Goyal, Piotr Bojanowski, and
  Armand Joulin.
\newblock Unsupervised learning of visual features by contrasting cluster
  assignments.
\newblock In {\em Adv. Neural Inform. Process. Syst.}, 2020.

\bibitem{simsiam21}
Xinlei Chen and Kaiming He.
\newblock Exploring simple siamese representation learning.
\newblock In {\em IEEE Conf. Comput. Vis. Pattern Recog.}, 2021.

\bibitem{xie2020propagate}
Zhenda Xie, Yutong Lin, Zheng Zhang, Yue Cao, Stephen Lin, and Han Hu.
\newblock Propagate yourself: Exploring pixel-level consistency for
  unsupervised visual representation learning.
\newblock In {\em IEEE Conf. Comput. Vis. Pattern Recog.}, 2021.

\bibitem{xie2021detco}
Enze Xie, Jian Ding, Wenhai Wang, Xiaohang Zhan, Hang Xu, Zhenguo Li, and Ping
  Luo.
\newblock Detco: Unsupervised contrastive learning for object detection.
\newblock In {\em Int. Conf. Comput. Vis.}, 2021.

\bibitem{xiao2021region}
Tete Xiao, Colorado~J Reed, Xiaolong Wang, Kurt Keutzer, and Trevor Darrell.
\newblock Region similarity representation learning.
\newblock In {\em Int. Conf. Comput. Vis.}, 2021.

\bibitem{Sivic2005DiscoveringOC}
Josef Sivic, Bryan~C. Russell, Alexei~A. Efros, Andrew Zisserman, and
  William~T. Freeman.
\newblock Discovering object categories in image collections.
\newblock In {\em Int. Conf. Comput. Vis.}, 2005.

\bibitem{RussellFESZ06}
Bryan~C. Russell, William~T. Freeman, Alexei~A. Efros, Josef Sivic, and Andrew
  Zisserman.
\newblock Using multiple segmentations to discover objects and their extent in
  image collections.
\newblock In {\em IEEE Conf. Comput. Vis. Pattern Recog.}, 2006.

\bibitem{KimT09}
Gunhee Kim and Antonio Torralba.
\newblock Unsupervised detection of regions of interest using iterative link
  analysis.
\newblock In {\em Adv. Neural Inform. Process. Syst.}, 2009.

\bibitem{FaktorI14}
Alon Faktor and Michal Irani.
\newblock ``clustering by composition'' - unsupervised discovery of image
  categories.
\newblock {\em IEEE Trans. Pattern Anal. Mach. Intell.}, 2014.

\bibitem{ChoKSP15}
Minsu Cho, Suha Kwak, Cordelia Schmid, and Jean Ponce.
\newblock Unsupervised object discovery and localization in the wild:
  Part-based matching with bottom-up region proposals.
\newblock In {\em IEEE Conf. Comput. Vis. Pattern Recog.}, 2015.

\bibitem{VoBCHLPP19}
Huy~V. Vo, Francis~R. Bach, Minsu Cho, Kai Han, Yann LeCun, Patrick
  P{\'{e}}rez, and Jean Ponce.
\newblock Unsupervised image matching and object discovery as optimization.
\newblock In {\em IEEE Conf. Comput. Vis. Pattern Recog.}, 2019.

\bibitem{vo2020toward}
Huy~V Vo, Patrick P{\'e}rez, and Jean Ponce.
\newblock Toward unsupervised, multi-object discovery in large-scale image
  collections.
\newblock In {\em Eur. Conf. Comput. Vis.}, 2020.

\bibitem{lod}
Huy~V. Vo, Elena Sizikova, Cordelia Schmid, Patrick P{\'{e}}rez, and Jean
  Ponce.
\newblock Large-scale unsupervised object discovery.
\newblock {\em arXiv: Comp. Res. Repository}, 2021.

\bibitem{JoulinBP10}
Armand Joulin, Francis~R. Bach, and Jean Ponce.
\newblock Discriminative clustering for image co-segmentation.
\newblock In {\em IEEE Conf. Comput. Vis. Pattern Recog.}, 2010.

\bibitem{HsuLC18}
Kuang{-}Jui Hsu, Yen{-}Yu Lin, and Yung{-}Yu Chuang.
\newblock Co-attention cnns for unsupervised object co-segmentation.
\newblock In {\em IJCAI}, 2018.

\bibitem{ChenL0H21}
Yun{-}Chun Chen, Yen{-}Yu Lin, Ming{-}Hsuan Yang, and Jia{-}Bin Huang.
\newblock Show, match and segment: Joint weakly supervised learning of semantic
  matching and object co-segmentation.
\newblock {\em IEEE Trans. Pattern Anal. Mach. Intell.}, 2021.

\bibitem{JiVH19}
Xu~Ji, Andrea Vedaldi, and Jo{\~{a}}o~F. Henriques.
\newblock Invariant information clustering for unsupervised image
  classification and segmentation.
\newblock In {\em Int. Conf. Comput. Vis.}, 2019.

\bibitem{HwangYSCYZC19}
Jyh{-}Jing Hwang, Stella~X. Yu, Jianbo Shi, Maxwell~D. Collins, Tien{-}Ju Yang,
  Xiao Zhang, and Liang{-}Chieh Chen.
\newblock Segsort: Segmentation by discriminative sorting of segments.
\newblock In {\em Int. Conf. Comput. Vis.}, 2019.

\bibitem{maskcontrast}
Wouter~Van Gansbeke, Simon Vandenhende, Stamatios Georgoulis, and Luc~Van Gool.
\newblock Unsupervised semantic segmentation by contrasting object mask
  proposals.
\newblock {\em arXiv: Comp. Res. Repository}, 2021.

\bibitem{resnet}
Kaiming He, Xiangyu Zhang, Shaoqing Ren, and Jian Sun.
\newblock Deep residual learning for image recognition.
\newblock In {\em IEEE Conf. Comput. Vis. Pattern Recog.}, 2016.

\bibitem{vnet}
Fausto Milletari, Nassir Navab, and Seyed-Ahmad Ahmadi.
\newblock V-net: Fully convolutional neural networks for volumetric medical
  image segmentation.
\newblock In {\em Proc.\ Int.\ Conf.\ 3D Vision}, 2016.

\bibitem{focalloss}
Tsung-Yi Lin, Priya Goyal, Ross Girshick, Kaiming He, and Piotr Doll{\'a}r.
\newblock Focal loss for dense object detection.
\newblock In {\em Int. Conf. Comput. Vis.}, 2017.

\bibitem{copypaste}
Golnaz Ghiasi, Yin Cui, Aravind Srinivas, Rui Qian, Tsung{-}Yi Lin, Ekin~D.
  Cubuk, Quoc~V. Le, and Barret Zoph.
\newblock Simple copy-paste is a strong data augmentation method for instance
  segmentation.
\newblock In {\em IEEE Conf. Comput. Vis. Pattern Recog.}, 2021.

\bibitem{coco}
Tsung{-}Yi Lin, Michael Maire, Serge~J. Belongie, James Hays, Pietro Perona,
  Deva Ramanan, Piotr Doll{\'{a}}r, and C.~Lawrence Zitnick.
\newblock Microsoft {COCO:} common objects in context.
\newblock In {\em Eur. Conf. Comput. Vis.}, 2014.

\bibitem{uvo}
Weiyao Wang, Matt Feiszli, Heng Wang, and Du~Tran.
\newblock Unidentified video objects: {A} benchmark for dense, open-world
  segmentation.
\newblock {\em arXiv: Comp. Res. Repository}, 2021.

\bibitem{voc}
Mark Everingham, Luc Van~Gool, Christopher~KI Williams, John Winn, and Andrew
  Zisserman.
\newblock {The PASCAL Visual Object Classes (VOC) Challenge}.
\newblock {\em Int. J. Comput. Vis.}, 2010.

\bibitem{mcg}
P.~Arbel\'{a}ez, J.~Pont-Tuset, J.~Barron, F.~Marques, and J.~Malik.
\newblock Multiscale combinatorial grouping.
\newblock In {\em IEEE Conf. Comput. Vis. Pattern Recog.}, 2014.

\bibitem{cob}
Kevis-Kokitsi Maninis, Jordi Pont-Tuset, Pablo Arbel\'{a}ez, and Luc~Van Gool.
\newblock Convolutional oriented boundaries: From image segmentation to
  high-level tasks.
\newblock {\em IEEE Trans. Pattern Anal. Mach. Intell.}, 2018.

\bibitem{MartinFTM01}
David~R. Martin, Charless~C. Fowlkes, Doron Tal, and Jitendra Malik.
\newblock A database of human segmented natural images and its application to
  evaluating segmentation algorithms and measuring ecological statistics.
\newblock In {\em Int. Conf. Comput. Vis.}, 2001.

\bibitem{MottaghiCLCLFUY14}
Roozbeh Mottaghi, Xianjie Chen, Xiaobai Liu, Nam{-}Gyu Cho, Seong{-}Whan Lee,
  Sanja Fidler, Raquel Urtasun, and Alan~L. Yuille.
\newblock The role of context for object detection and semantic segmentation in
  the wild.
\newblock In {\em IEEE Conf. Comput. Vis. Pattern Recog.}, 2014.

\bibitem{lvis2019}
Agrim Gupta, Piotr Dollar, and Ross Girshick.
\newblock {LVIS}: A dataset for large vocabulary instance segmentation.
\newblock In {\em IEEE Conf. Comput. Vis. Pattern Recog.}, 2019.

\bibitem{lost}
Oriane Sim{\'{e}}oni, Gilles Puy, Huy~V. Vo, Simon Roburin, Spyros Gidaris,
  Andrei Bursuc, Patrick P{\'{e}}rez, Renaud Marlet, and Jean Ponce.
\newblock Localizing objects with self-supervised transformers and no labels.
\newblock {\em arXiv: Comp. Res. Repository}, 2021.

\bibitem{selectivesearch}
Jasper R.~R. Uijlings, Koen E.~A. van~de Sande, Theo Gevers, and Arnold W.~M.
  Smeulders.
\newblock Selective search for object recognition.
\newblock {\em Int. J. Comput. Vis.}, 2013.

\bibitem{bar2021detreg}
Amir Bar, Xin Wang, Vadim Kantorov, Colorado~J Reed, Roei Herzig, Gal Chechik,
  Anna Rohrbach, Trevor Darrell, and Amir Globerson.
\newblock Detreg: Unsupervised pretraining with region priors for object
  detection.
\newblock {\em arXiv: Comp. Res. Repository}, 2021.

\bibitem{wei2019unsupervised}
Xiu-Shen Wei, Chen-Lin Zhang, Jianxin Wu, Chunhua Shen, and Zhi-Hua Zhou.
\newblock Unsupervised object discovery and co-localization by deep descriptor
  transformation.
\newblock {\em Pattern Recognition}, 2019.

\bibitem{mocov2}
Xinlei Chen, Haoqi Fan, Ross Girshick, and Kaiming He.
\newblock Improved baselines with momentum contrastive learning.
\newblock {\em arXiv: Comp. Res. Repository}, 2020.

\bibitem{dino}
Mathilde Caron, Hugo Touvron, Ishan Misra, Herv\'e J\'egou, Julien Mairal,
  Piotr Bojanowski, and Armand Joulin.
\newblock Emerging properties in self-supervised vision transformers.
\newblock In {\em Int. Conf. Comput. Vis.}, 2021.

\bibitem{li2021esvit}
Chunyuan Li, Jianwei Yang, Pengchuan Zhang, Mei Gao, Bin Xiao, Xiyang Dai,
  Lu~Yuan, and Jianfeng Gao.
\newblock Efficient self-supervised vision transformers for representation
  learning.
\newblock {\em arXiv: Comp. Res. Repository}, 2021.

\bibitem{fpn}
Tsung{-}Yi Lin, Piotr Doll{\'{a}}r, Ross~B. Girshick, Kaiming He, Bharath
  Hariharan, and Serge~J. Belongie.
\newblock Feature pyramid networks for object detection.
\newblock In {\em IEEE Conf. Comput. Vis. Pattern Recog.}, 2017.

\end{thebibliography}
}

\clearpage
\newpage
\appendix

\section*{Appendix}

\renewcommand{\thefigure}{S\arabic{figure}}
\setcounter{figure}{0}
\renewcommand{\thetable}{S\arabic{table}}
\setcounter{table}{0}

We provide more information here.

\section{Additional implementation details}\label{app:impl}

\subsection{Evaluation protocol}\label{app:eval}
While evaluating the performance of class-agnostic instance segmentation, we also report the results of an easier protocol  AP$^*$ which  evaluates medium and large objects.
Here, only the objects with area greater than $64^2$ are considered and their mask AP$^*$ with an IoU threshold of $0.5$ is computed.
AP$^*_M$ and AP$^*_L$ are also reported for medium and large objects, 
\ie, objects with area in the range of $(64^2, 192^2)$  and those with area greater than $192^2$, respectively.
The results of MCG and COB are computed using the official segmentation masks.

\subsection{Supervised fine-tuning}\label{app:ft}
We evaluate the pre-trained instance segmentation model by fine-tuning it with manual annotations.
Specifically, we fine-tune a dynamic SOLO model (aka SOLOv2) on COCO \texttt{train2017} and evaluate on COCO \texttt{val2017}.
Synchronized batch normalization is used in the backbone along with FPN~\cite{fpn} during training.
We provide two training settings, \ie, limited fully annotated images, and limited segmentation masks. 

\myparagraph{Limited images.}
For the experiments with limited images, we use 5\% and 10\% images from COCO \texttt{train2017}, which corresponds to \app6k and \app12k fully annotated images, respectively.
We  fine-tune the instance segmenter initialized with the pre-trained model for 20k iterations with an initial learning rate of 0.01, which is then divided by 10 at 12k and 18k iterations.

\myparagraph{Limited masks.}
For the experiments with limited masks, we use 5\% and 10\% segmentation masks from COCO \texttt{train2017}.
In this setting, only 5\% and 10\% of the images have mask annotations, \ie, \app6k and \app12k images, respectively. 
Specifically, we use all the class labels to supervise the category branch, but only use a part of the annotated masks to supervise the mask branch. The model is trained for 90k iterations with the standard schedule. 

\subsection{Training details}
For the self-supervised pre-trained backbones, we use the official models trained on ImageNet without labels for 200 epochs.
For \Ours, we use the images in COCO \texttt{train2017} and COCO \texttt{unlabeled2017} as the set of unlabeled images, containing a total of $\app$241k images.
We use ResNet-50 as the backbone for all the fine-tuning experiments and ablation study and use ResNet-101 for other results and visualizations.
We train for 30k iterations on 8 GPUs with a total of 32 images per mini-batch. The learning rate is set to 0.0025.
In the self-training, we repeat the schedule once and train for another 30k iterations.

\myparagraph{Copy-paste augmentation.}
For a pair of images in a batch, we randomly select objects from one image and paste them at random locations on the other image. These objects are not pasted if they have a high overlap (IoU $>= 0.5$) with existing objects.

\section{Additional results}
We report the results of an easier protocol  AP$^*$ which  evaluates medium and large objects in Table~\ref{tab:rebuttal_ar_insseg}.
As shown, the gains over MCG and COB are larger, especially for the large objects.

\iftrue
\begin{table}[h]
\small
\tablestyle{5pt}{1.1}
\centering
\resizebox{0.87\columnwidth}{!}{
\begin{tabular}{l|ccc|ccc}
\toprule
 method  &  AP$_\text{50}$ & AP$_\text{75}$ & AP & AP$^{*}$  &  AP$^*_{M}$ & AP$^*_{L}$ \\ 
\midrule
\emph{w/ anns:} &&&\\
MCG~\cite{mcg}  &	4.6 &	0.8 & 1.6  & 9.4 &	32.4 &	7.4  \\
COB~\cite{cob}  &	8.8 &	1.9 & 3.3 & 15.6 &	36.5 &	11.0  \\
\midrule
\emph{w/o anns:} &&&\\
\textbf{\Ours}   &  9.8  & 2.9 & 4.0   & 24.3  &  21.5  & 34.3 \\
\bottomrule
\end{tabular}
}
\caption{Class-agnostic instance segmentation on COCO \texttt{val2017}. 
Both MCG and COB require annotations. 
}
\label{tab:rebuttal_ar_insseg}
\end{table}
\fi

\section{Additional visualizations}

\begin{figure*}[t!]
\centering
   \includegraphics[width=1.0\linewidth]{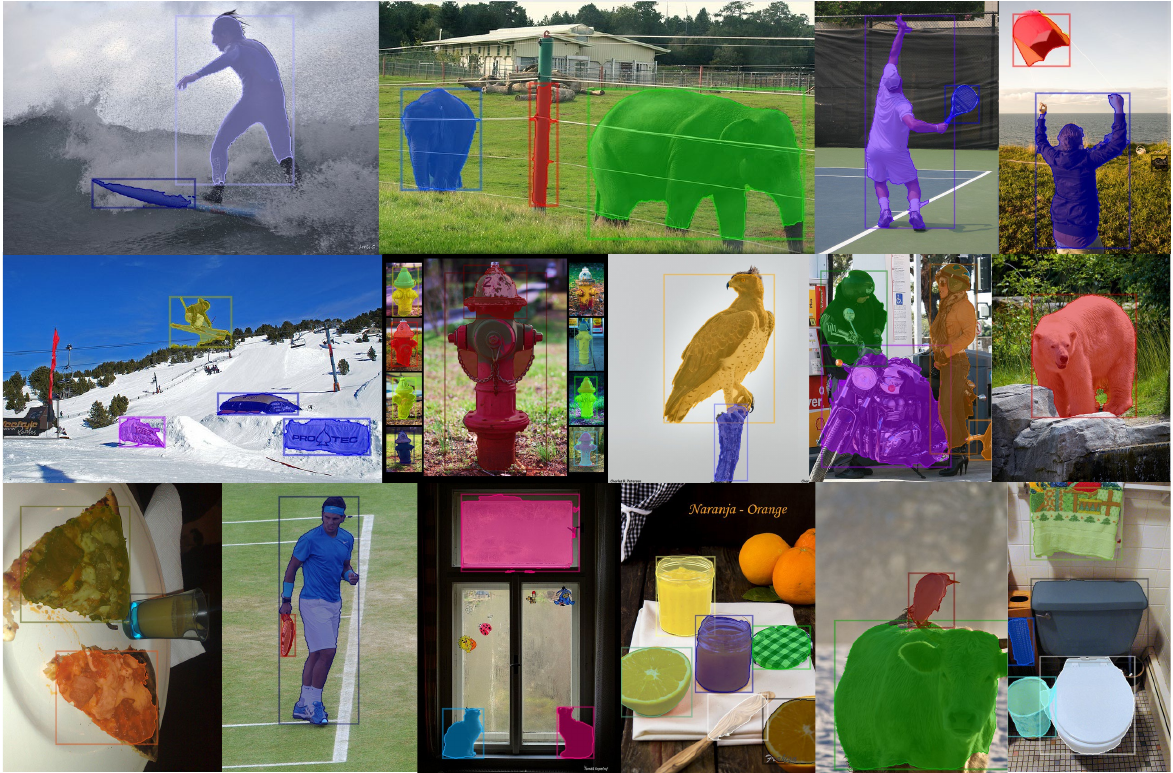}
    \vspace{-0.55cm}
    \captionof{figure}{%
    \textbf{More qualitative results of \Ours
    for the task of class-agnostic instance segmentation.}
    The model is trained \textit{without any kind of manual annotations} and can infer at 16 FPS on a V100 GPU.  
    Best viewed on screen.
    }
\label{fig:app_vis}
\vspace{-0.5em}
\end{figure*}

\begin{figure}[t!]
\centering
\includegraphics[width=0.76\linewidth]{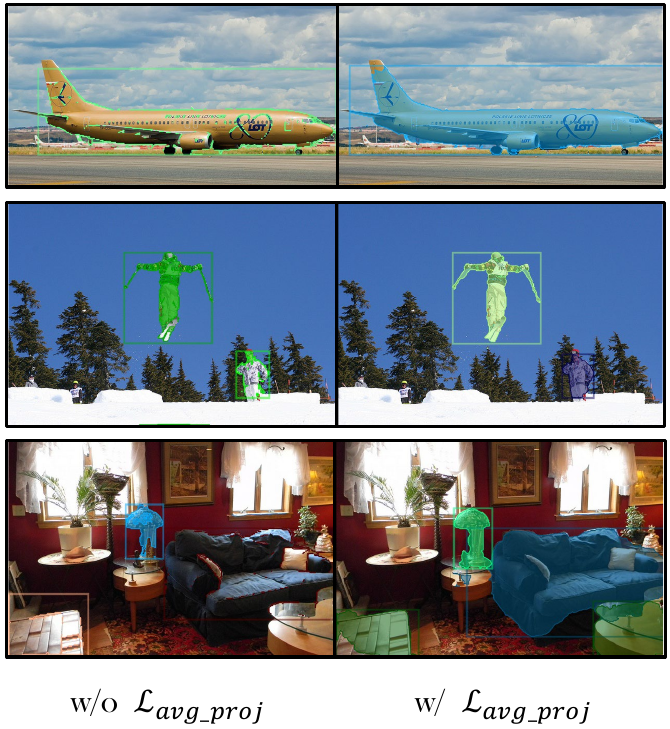}
\vspace{-0.5em}
\caption{
Qualitative comparison of \Ours with and without $\mathcal{L}_{avg\_proj}$ when learning from coarse masks.
The model trained without $\mathcal{L}_{avg\_proj}$ tends to only segment the contours when trained longer. %
}
\label{fig:app_avgproj_vis}
\vspace{-0.5em}
\end{figure}

\begin{figure}[t!]
\centering
\includegraphics[width=0.99\linewidth]{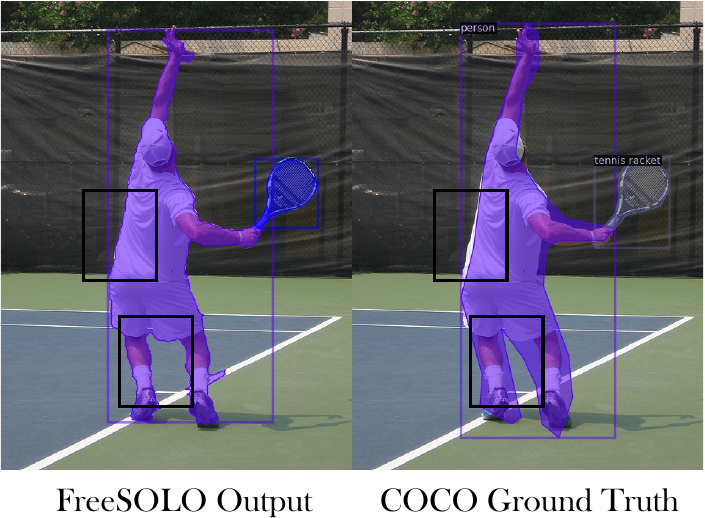}
\vspace{-0.5em}
\caption{
Qualitative comparison of \Ours's predicted masks and ground truth masks. At some object boundaries, \Ours can  produce
\textit{even
more precise} segmentation than manual annotations in some cases.
}
\label{fig:app_comparision_gt}
\vspace{-0.5em}
\end{figure}

In this section, we provide additional visualizations of \Ours.
We show qualitative results of our method for the task of class-agnostic instance segmentation in Figure~\ref{fig:app_vis}.
In Figure~\ref{fig:app_avgproj_vis}, we provide more qualitative comparison of \Ours with and without the $\mathcal{L}_{avg\_proj}$.
As shown in Figure~\ref{fig:app_comparision_gt}, we further show that \Ours can even produce more precise segmentation results than manual annotations at some object boundaries, which indicates \Ours's great potential for tasks such as auto-labeling.

\end{document}